# Learning Dynamic and Hierarchical Traffic Spatiotemporal Features with Transformer

Haoyang Yan, Xiaolei Ma

## Abstract

Traffic forecasting is an indispensable part of Intelligent transportation systems (ITS), and long-term network-wide accurate traffic speed forecasting is one of the most challenging tasks. Recently, deep learning methods have become popular in this domain. As traffic data are physically associated with road networks, most proposed models treat it as a spatiotemporal graph modeling problem and use Graph Convolution Network (GCN) based methods. These GCN-based models highly depend on a predefined and fixed adjacent matrix to reflect the spatial dependency. However, the predefined fixed adjacent matrix is limited in reflecting the actual dependence of traffic flow. This paper proposes a novel model, Traffic Transformer, for spatial-temporal graph modeling and long-term traffic forecasting to overcome these limitations. Transformer is the most popular framework in Natural Language Processing (NLP). And by adapting it to the spatiotemporal problem, Traffic Transformer hierarchically extracts spatiotemporal features through data dynamically by multi-head attention and masked multi-head attention mechanism, and fuse these features for traffic forecasting. Furthermore, analyzing the attention weight matrixes can find the influential part of road networks, allowing us to learn the traffic networks better. Experimental results on the public traffic network datasets and real-world traffic network datasets generated by ourselves demonstrate our proposed model achieves better performance than the state-of-the-art ones.

## 1 Introduction

Urban transportation systems are satisfying travelers' demands and ensure the operation of cities. But with the urbanization process continuing to accelerate, many problems such as traffic congestion are getting tough to handle. Intelligent transportation systems (ITS) can help alleviate traffic congestion, and traffic speed prediction is regarded as the ITS's foundation. Traffic speed is one of the most critical indicators to represent the traffic conditions. Numbers of researchers proposed different methods aiming at traffic speed forecasting, and achieved great success.

Recently, deep learning methods have become popular in this domain, which allows researchers to build data-driven models to extract the spatiotemporal features. The network wide traffic flow data has two sorts of features: spatial feature and temporal feature. Temporal feature can be extracted though different time-series models, but spatial feature still hasn’t a general way to extract. In the beginning, researchers ignored the spatial feature, and only focus on single roads. Former deep learning models usually transformed traffic network to grid-like structure, and treated the network like an image to apply Convolution Neural Network (CNN) extracting spatial feature. However, these transforms obviously lost amounts of information and break the network relationship completely. For example, roads in the same grid are merged and cannot be analyzed. As traffic data are physically associated with road networks, recently proposed models treat it as a spatiotemporal graph modeling problem and use Graph Convolution Network (GCN) based methods. The key of these GCN-based models is the adjacent matrix. The adjacent matrix defines the relationship of nodes and edges of the graph that allows model convolutional gathering the information in neighborhood nodes. GCN-based models have recently achieved tremendous success in network-wide traffic forecasting. However, there are still some limitations in these GCN-based models.

### *1.1 Limitations in GCN-based models*

a) Defining a perfect adjacent matrix is difficult and costly for humans. Adjacent matrix represents the message passing in graph, but the propagation of traffic is complex, not simply based on distance, and some nodes can be regarded as abstract connections. For example, a sensor in the downtown junction can represent the level of service of the network to a certain extent. This sensor is influential and can be regarded as connected to most of the sensors abstractly.
b) Adjacent matrix is fixed in models. Fixed adjacent matrix cannot handle the dynamic of traffic in different situations, for example, week and weekend, morning peak and evening peak. Besides, as we conclude in (a), there are inevitable mistakes in adjacent matrix. For example, in a No left-turn intersection, the sensors upstream and downstream of the left-turn route are spatial close, but they are far in logic. Fixed adjacent matrix means these mistakes affect the model all the way.
c) Not deep and hierarchical. Proposed models usually just use a single adjacent matrix. Some models use a set of adjacent matrixes, but the results of these adjacent matrixes are just simply added or concatenated. However, we always hope a hierarchical, “deep” model in deep learning.

Aiming at tackling the challenges mentioned above, we find spatial feature extraction and natural language processing (NLP) are similar, and learn from the advanced methods of NLP.

### *1.2 Similarity between spatial feature extraction and NLP*

Spatial feature extraction and NLP are similar not only in data characteristics but also in challenges and difficulties. As shown in Fig.1 (a), both natural language data and traffic data can be regarded as sequential data. Natural language data is the sequence of words and traffic data is the sequence of sensors. As shown in Fig.1 (b), the absolute and relative position is important. Changing the order of the sequence will change the meaning. As shown in Fig.1 (c), long-term dependency is usual in NLP, deducing the date of tomorrow must back to the beginning of the article. Each sensor in traffic network may influent sensors abstractly no matter far or close. As shown in Fig.1 (d), the dependency is dynamic, the meaning of word depends on the context. Even if the words “it” in two sentences are exactly the same, the meaning of two words is different. Similarly, the traffic state in the center nodes are same, but the spatial dependencies of the nodes are different.

Transformer achieves great successes in NLP by dealing the challenges and difficulties above, and Transformer is a robust framework for sequence learning. Consider the similarity, this paper proposes a novel model, Traffic Transformer, for spatial-temporal graph modeling and long-term traffic forecasting to overcome the limitations in GCN-based models. We adapt Transformer to traffic forecasting problems. By the multi-head attention mechanism of Transformer, each pair of nodes' relationships are extracted dynamically through data. Stacking Transformer layers allow our model to hierarchically extract features. And hierarchically scaled features are fused by attention mechanism. Experimental results on the public traffic network datasets, METR-LA, and two real-world traffic network datasets generated by ourselves, demonstrate our proposed model achieves better performance than the state-of-the-art ones. The main contributions of our work are summarized as follows:

- We present a novel model called Traffic Transformer for spatial-temporal graph modeling and long-term traffic forecasting. It extracts and fuses both global and local spatial features of network-wide traffic flow, and significantly improves prediction accuracy, especially in long-term prediction.
- Extracted spatial dependencies are dynamic in Traffic Transformer depending on input data. For different situations of different input data, Traffic Transformer gives different spatial relationships to give a better prediction by attention mechanism.
- Features extracted hierarchically in Traffic Transformer. Different layers and blocks learn different features and fuse hierarchically, and help us better learn the traffic network and the propagation of traffic flow.
- Experimental results show even the adjacent matrix is unknown to our model, the model can still perform well-enough and make good prediction.

The rest of this paper is organized as follows: Section 2 present a literature review of traffic speed prediction methods. Section 3 shows the formulation of the traffic forecasting problem and our proposed Traffic Transformer network. Section 4 provides experiments to evaluate our model's performance and the interpretation of the dynamic and hierarchical traffic spatiotemporal features extracted. Finally, we conclude our paper and present our further work in Section 5.

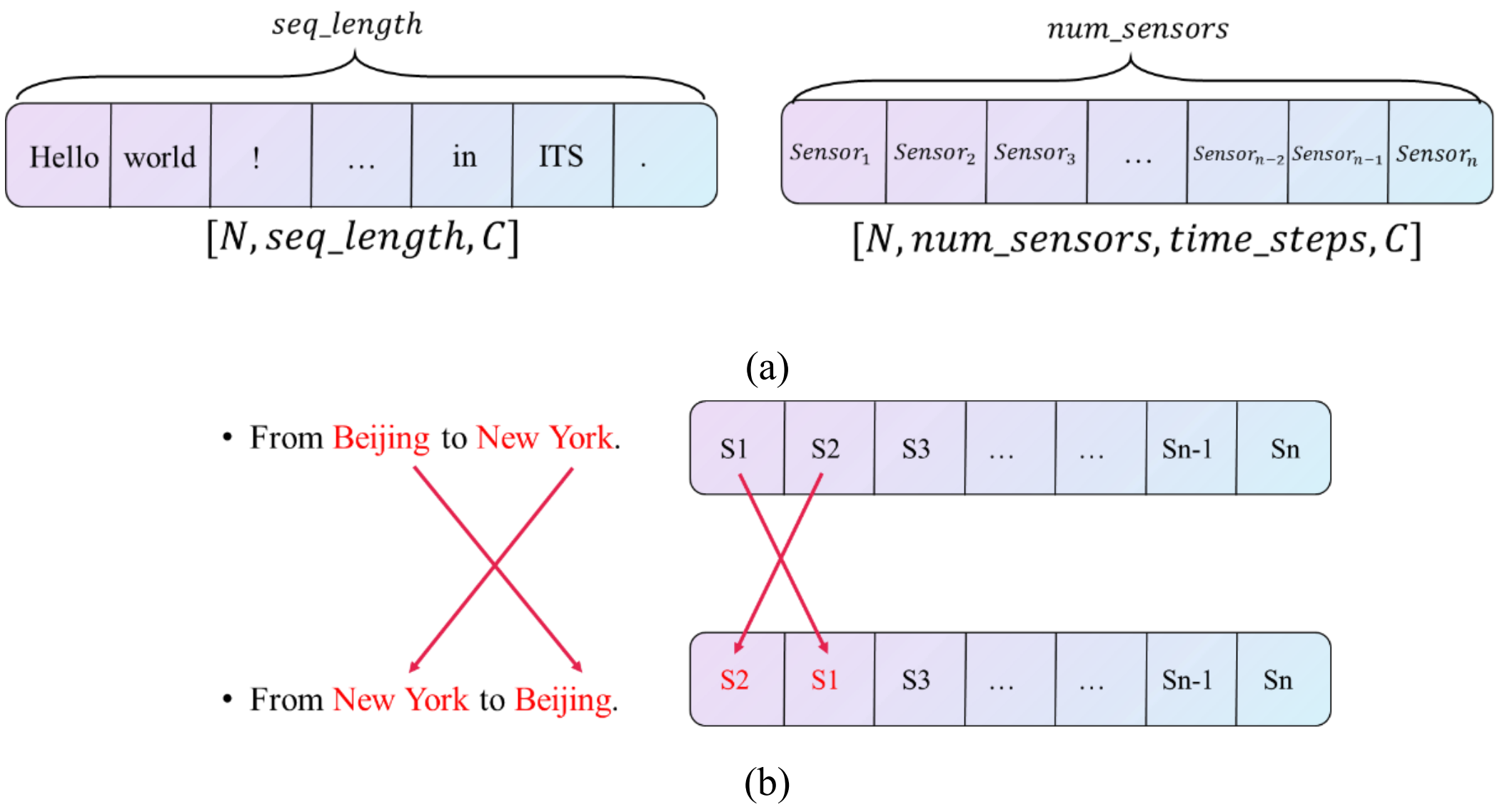

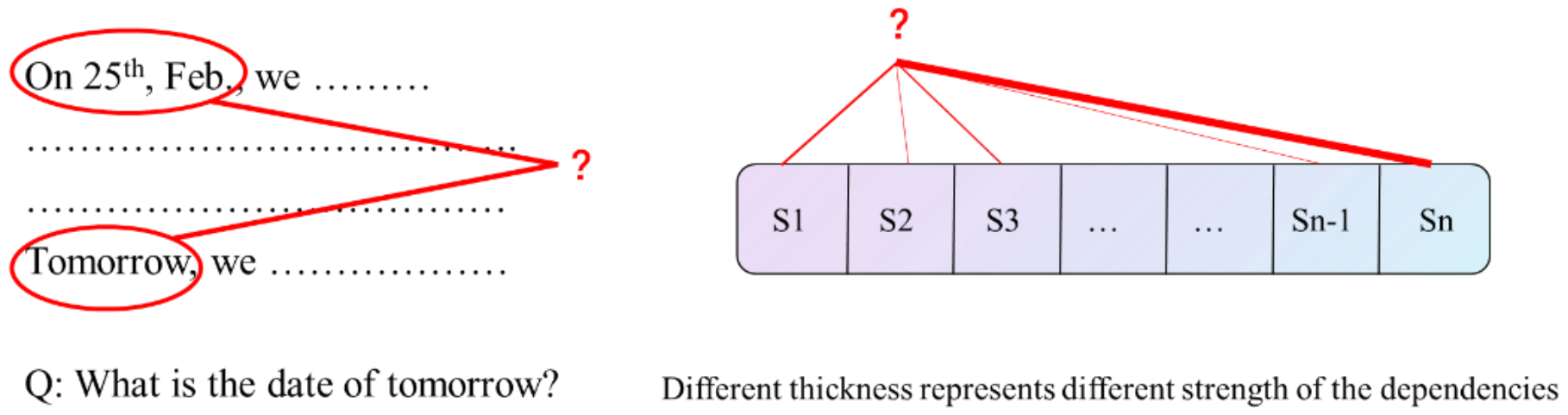

(c)

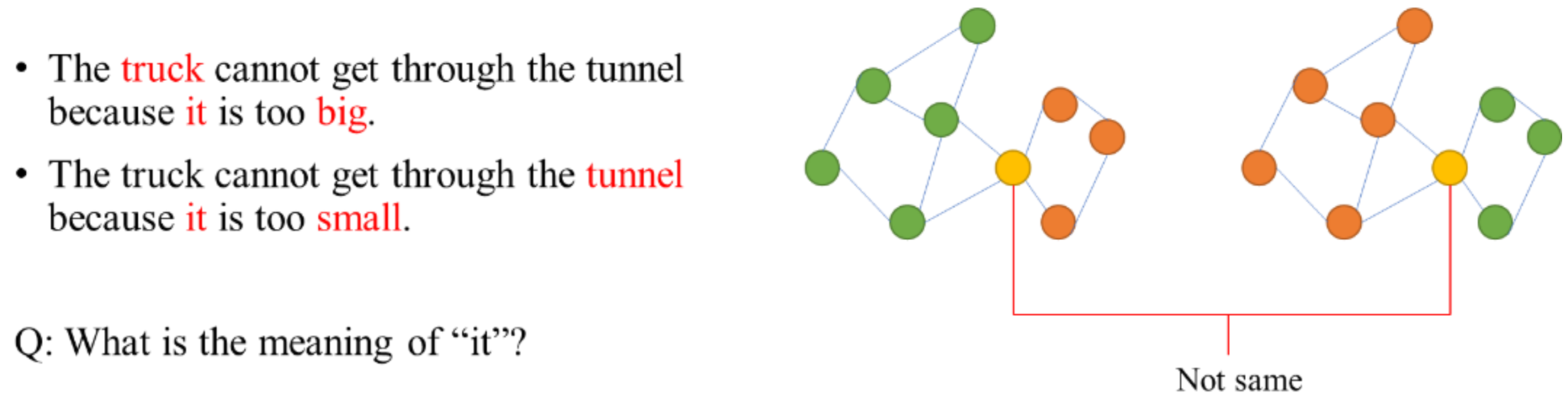

(d)

**Figure1: The similarity between spatial feature extraction and NLP. (a): Natural language data and traffic data can be regarded as sequential data; (b): Changing the order of the sequence will change the meaning; (c): High dependencies may exist even far away in the sequence; (d): The meaning is dynamic and depends on other parts.**

## 2 Literature Review

Over the past few decades, researchers proposed numbers of different approaches for traffic forecasting. The literatures in this filed are aiming to better accuracy. To achieve this goal, existing approaches are extracting and modeling the spatial and temporal relationships of traffic data. We divided the former research into three different stages: (a) Traditional approaches; (b) Grid-based deep learning; (c) Graph-based deep learning. And this division is depending on how spatial features is extracted. The brief literature review of these three stages is summarized below. And as we mentioned above, spatial feature extraction and NLP are similar not only in data characteristics but also in challenges and difficulties. We are adapting the advanced methods of NLP to traffic forecasting. Therefore, the brief literature review of sequence modeling in NLP is also summarized as follows.

### *2.1 Traditional approaches*

As early as 1976, AutoRegressive Integrated Moving Average (ARIMA) was propsed by Box and Jenkins. In 1979, ARIMA was applied in traffic forecast (Ahmed and Cook, 1979). ARIMA is a parametric method, which assumes that the distribution of traffic state is known and can be estimated by several parameters. Variants of ARIMA show better performance by improving ARIMA to consider more influence in traffic system. For example, Seasonal ARIMA (SARIMA) (Williams & Hoel, 2003) considers the periodic effects in the transportation system, KARIMA (Van Der Voort, Dougherty, and Watson 1996) combined kohonen maps with ARIMA for traffic forecasting, Lee and Fambro used subset ARIMA for short-term freeway traffic volume forecasting. Other parametric methods including exponential smoothing (ES) (Ross, 1982) and Kalman filter (Okutani and Stephanedes, 1984). However, the distribution of traffic state is complex and dynamic, these parametric methods may not perform well to the sudden changes in traffic system such as peak hours.

Some advanced none-parametric machine learning methods are proposed for better accuracy. $k$-NN (Davis and Nihan 1991; Cai et al., 2016) is non-parametric lazy model that predict future state by finding the $k$ nearest neighborhoods of input. Bayesian network (Sun, Zhang, and Yu 2006) uses causal network to represent the probability of variables. Support vector machine (SVM) (Vanajakshi and Rilett, 2004; Zhang and Xie, 2008) defines the support vector and make prediction though support vector. Random forests (RF) (Leshem and Ritov, 2007) and gradient boosting decision tree (GBDT) (Zhang and Haghani, 2015) are ensemble method that bagging or boosting the decision tree. Wavelet packet-autocorrelation function (Jiang and Adeli, 2004) uses discrete wavelet packet transform to representing additional subtle details of a signal, Artificial Neural Networks (ANN) (Adeli, 2001; Dharia and Adeli, 2003; Vlahogianni et al., 2007) imitates biological neuron to construct neural network building the relationship between input and output. However, these methods are focus on short-term traffic forecasting in regular period. Due to the high dynamics and non-linearity within traffic flow, these models not perform well to peak hours and in long-term traffic forecasting.

### *2.2 Grid-based deep learning*

Recent years, researchers pay much attention to deep learning models on account of their ability of extracting the nonlinearity and features. The deep neural networks (DNN) can extract the spatial and temporal features hidden in traffic flow data, and significantly improve models' performance. Many literatures proposed different DNN based models, and deep learning methods have been widely used in the field of civil engineering (Rafiei and Adeli,2016, 2018; Rafiei, Khushefati, Demirboga, and Adeli,2017). DNN models can be deployed for both spatial and temporal features extracting. For example, Ma et al. (2015) uses long-short term memory (LSTM) networks to predict speed evolution on a corridor. M. Zhou et al. (2017) proposed a recurrent neural network (RNN)-based microscopic car-following model to predict future traffic oscillations. Ma et al. (2017) converts traffic dynamics to heat map images and employs deep convolutional neural networks (CNN) for speed prediction. H. Yu et.al. proposed a spatiotemporal recurrent convolutional network for traffic prediction (2017). A hybrid CNN-LSTM algorithm developed by S. Chen, Leng, and Labi (2019) considered both human prior knowledge and time information. M. Zhou, Yu and Qu (2020) combined reinforcement learning and the car-following model to improve the driving strategy for connected and automated vehicles. These DNNs achieve much accuracy and robust in features extraction and forecasting. However, the input data of these methods are limited in Euclidean structures. Network-wide traffic data is kind of spatiotemporal data. The temporal dimension is naturally in Euclidean structures, but the spatial dimension is not. Although the traffic data are naturally connected to the graph domains, these methods need to transform the graph data to grid-like Euclidean structures, which lost lots of information in transformation. For example, nodes in the same grid are consider same and their information are lost. Besides, granularity is a hyperparameter of these "graph to grid" transform methods which is usually hard to define.

### 2.3 Graph-based deep learning

As traffic dynamics naturally connected with the road network, we hope we can analysis the problems directly on graph. The convolutional operator is extended to graph convolution (Kipf and Welling, 2017), and called Graph Convolution Networks (GCN). GCN can better extract the network-wide spatial features as the traffic data are naturally represented by a graph. For example, Li et al. (2018) proposed Diffusion Convolutional Recurrent Neural Network (DCRNN), which combined diffusion convolution and gated recurrent units (GRU). Yu et al. (2018) combined spatial graph convolution and temporal gated convolution to network-wide traffic forecasting, so-called spatiotemporal graph convolutional networks (STGCN). Yu et al. (2019) combined U-Net and STGCN to extract spatiotemporal features in hierarchical levels, so-called ST-UNet. These models have a limitation that the spatial dependences of these models are predefined and stay fixed after training. Usually the distance and up-down stream relationship are considered and finally an adjacent matrix will be calculated and implement into models. To deal with the above limitation, new models are proposed to learn spatial correlations from data. Zhang et al. (2018) proposed gated attention networks (GaAN) learn dynamic spatial correlations from data by graph attention (GAT) mechanism (Veličković et al., 2018). Wu et al. proposed Graph Wave Net (GWN) and using source vector and target vector to embedding and learn the adjacent matrix from data. The success of these "adapting adjacent matrix" method indicate that the spatial dependencies of traffic are not simply limited on road networks, long-distance spatial dependencies exist, and the nodes in the junction can be seen as connecting to the whole network. However, although the adjacent matrixes in these methods are learnable, the adjacent matrixes are still fixed after training.

### 2.4 Sequence modeling in NLP

NLP is always one of the most popular field in artificial intelligence. Most data in NLP are sequential and sequence modeling is the foundation of NLP. Sequence to sequence (seq2seq) is one of the basic tasks of NLP, seq2seq2 means that both the input and output of the model are sequence that the length of the sequence is uncertain. For example, the input of machine translation is sentences that the length is uncertain, and the output of machine translation is also the sentences that the length is uncertain. Similar tasks include question answering, text generation, etc. Sutskever et.al. proposed a seq2seq model with neural network in 2014 that encode the input to a vector and decoder the vector to output. Attention mechanism is proposed and combined with seq2seq models and shows tremendous performance. Attention mechanism can map a query vector and a set of key-value pairs vectors to an output. The output is computed as a weighted sum of the values, where the weight assigned to each value is calculated by a compatibility function of the query with the corresponding key. Attention mechanism allows models back to the input sequence of works and find the important part while outputting. Attention mechanism now is widely used in most deep learning tasks including traffic forecasting. For example, Q. Liu et.al proposed a short-term traffic speed forecasting method based on attention convolutional neural network.

In a long period, researchers believed CNN or RNN are indispensable part in sequence modeling which can handle the uncertain length with in the sequence data, but Vaswani et al. (2017) proposed Transformer which completely abandoned the CNN and RNN. Transformer only used attention mechanism and fully-connected forward neural network to model sequences. Transformer and its variants achieve most of the state-of-the-art performances. It indicates Transformer has powerful ability in sequence and non-local modeling, which is much suitable in extracting spatial features of traffic network. Recently, numbers of researchers proposed different models based on the idea of Transformer to adapt Transformer for tasks in other field including Computer Version (Dosovitskiy et.al, 2021) and Point Could (M. Guo et.al, 2021), and these variants achieved state-of-the-art performance in their field. It means that Transformer is transferable and can be commonly used in many fields including traffic forecasting in transportation. However, there is few traffic forecasting models consider using Transformer.

## 3 Methodology

This section presents the Traffic Transformer framework, which adapts Transformer for network-wide traffic speed forecasting. The model not only shows better than state-of-the-art performance but also has excellent interpretability to understanding hierarchical traffic spatiotemporal features.

### *3.1 Problem statement*

Network-wide traffic dynamics can naturally be written as a spatiotemporal graph:

$$\mathcal{G}(\mathcal{V}_N, \mathcal{E}, \boldsymbol{A}_{N\times N}; \boldsymbol{X}_{T\times N\times C})$$

where $\mathcal{V}_N$ is the set of $N$ nodes that represent the sensors, $\mathcal{E}$ is the set of edges that connect these nodes, and $\boldsymbol{A}_{N\times N}$ is the adjacent matrix which is usually pre-constructed based on the distances and up-down stream relationship between nodes. The evolution of traffic states in $T$ discrete time steps of duration $\Delta t$ is represented by the feature tensor $\boldsymbol{X}$. The feature vector of node $i$ at timestep $t$ is denoted as $X_i^t \in \mathbb{R}^C$ which is composed of $C$ traffic states such as speed, time of day, day of the week, etc. The traffic forecasting problem is a classic spatiotemporal prediction problem. Formally, the input is the feature tensor in the past $m$ timesteps $\boldsymbol{X}_{m\times N\times C}$ on $\mathcal{G}$, the output is the predicted feature tensor of the next $n$ timesteps $\boldsymbol{X}_{n\times N\times C}$. So the network-wide traffic forecasting problem can generally be formulated as

$$\boldsymbol{X}_{n\times N\times C} = \mathcal{F}(\boldsymbol{X}_{m\times N\times C}; \mathcal{G})$$

where $\mathcal{F}$ is the model to learn. In this paper, we apply Traffic Transformer to learn the hierarchical traffic spatiotemporal features and achieve network-wide traffic forecasting.

### *3.2 The Overall Architectures*

As shown in Fig.2, the Traffic Transformer consists of two main parts. One is called Global Encoder, and another is called Global-Local Decoder. Several Global Encoder and Global-Local Decoder blocks are stacked to form a deep model for hierarchical features. Global Encoder and Global-Local Decoder extract global spatial features and local spatial features respectively. And Global-Local Decoder also fuses the global spatial features extracted by Global Encoder and local spatial features extracted by Global-Local Decoder. Besides, the Temporal Embedding block extracts the temporal features at the beginning of the model. Then, the Positional Encoding and Embedding block helps model understanding the absolute and relative position of nodes. Finally, a dense neural network neural network aggregates the learned features for final predictions. There are usually two different ways to train the model. Previous models usually think the traffic forecasting as an auto-regressive problem and make predictions step-by-step, which causes error accumulation problems. So, our model abandons the auto-regressive method, and multi-step predictions are made at the same time. In this way, the accuracy of long-term prediction is improving and the inferring time is shrinking.

The key of Traffic Transformer is the Multi-head attention block. Multi-head attention is one of the self-attention mechanisms. And self-attention is a powerful way for "non-local" features extraction. Multi-head attention can find the relationships between every pair of nodes no matter far or close, and calculate the different weight of each relationship. The output of multi-head attention depends on every node's value and weight. To be point out, the weight of attention can be formed as an attention weight matrix that seems like an adjacent matrix. So, Multi-head attention can be seen as a GNN-based module. Still, the adjacent matrix is learned from data, and the adjacent matrix dynamically changes depending on the input data. The Multi-head attention in Global Encoder extracts the global dependencies. The Masked Multi-head attention in Global-Local Decoder is a variant of Multi-head attention that uses a mask to ignore some nodes to focus on extracting local features. The Multi-head attention in Global-Local Decoder fuses both global and local features of each node. We will discuss more details of Multi-head attention in the sections below.

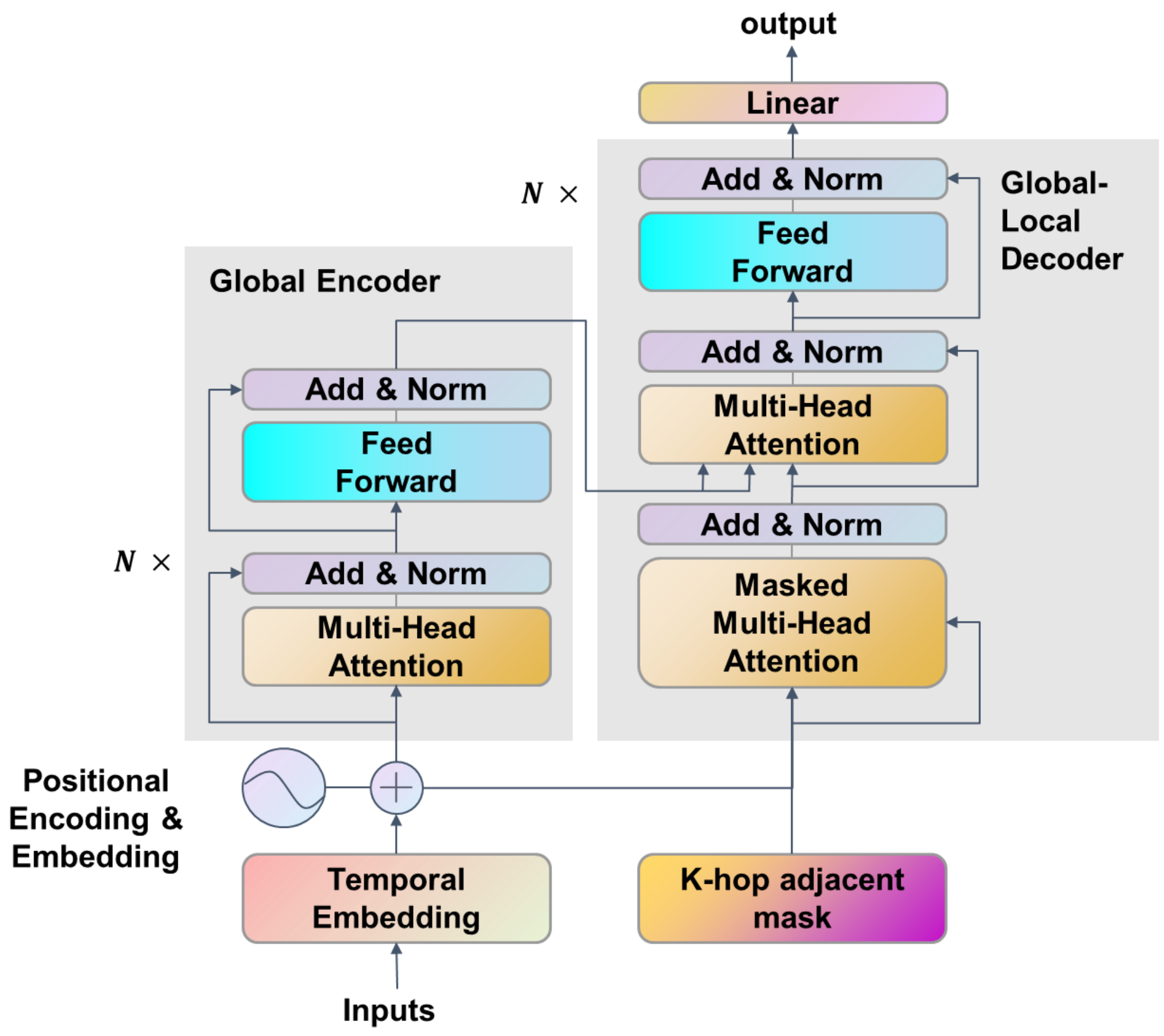

**Figure 2: Traffic Transformer's Architecture**

*3.3 Global Encoder*

This subsection concentrates on the global spatial features that the features between each node and all other nodes. Each Global Encoder block has two parts. The first is a Multi-head attention blcok, and the second is a fully connected feed-forward layer. The residual connection is added around each sub-layer followed by layer normalization to stabilize the gradient and help training the model better. We will further elaborate on how global spatial features are extracted by explaining the parts below.

1) Multi-head attention

An attention function can map a query vector and a set of key-value pairs vectors to an output. The output is computed as a weighted sum of the values, where the weight assigned to each value is calculated by a compatibility function of the query with the corresponding key. Scaled Dot-Product Attention is commonly used in Transformer, which can be formulated as

$$Attention(Q,K,V) = softmax\left(\frac{QK^T}{\sqrt{d_k}}\right)V$$

where a set of queries packed together into a matrix $Q$, and the keys and values are packed into matrix $K, V$ respectively. $d_k$ is the dimension of queries and keys.

In Global Encoder, matrix $Q, K, V$ are derived from the same input features. The input features are first projected to different latent subspace with different learnable feed-forward neural networks. Which can be formulated as

$$\begin{cases} Q = XW_q \\ K = XW_k \\ V = XW_v \end{cases}$$

where $X$ is the input features, and $W_q, W_k, W_v$ represent the learnable parameters of different feed-forward neural networks respectively.

Besides, we define the attention weight matrix $\boldsymbol{A} \in \mathbb{R}^{N\times N}$ as

$$\boldsymbol{A} = softmax\left(\frac{QK^T}{\sqrt{d_k}}\right)$$

As shown in Fig.3, the attention weight matrix can represent the dependencies between every pair of nodes. And the attention weight matrix dynamically changes depending on the input data.

Multi-head attention uses different learned feed-forward neural networks to linearly project the $Q, K, V$ for $h$ times. It allows the model to jointly attend to information from different representation subspace at different positions. And it can be formulated as

$$MultiHead(Q,K,V) = Concat(head1, \dots, head_h)W^O$$

$$where\ head_i = Attention(QW_i^Q, KW_i^K, VW_i^V)$$

Where the projections are parameter matrices $W_i^Q \in \mathbb{R}^{d_{model}\times d_k}, W_i^K \in \mathbb{R}^{d_{model}\times d_k}, W_i^V \in \mathbb{R}^{d_{model}\times d_v}$, $W^O \in \mathbb{R}^{hd_v\times d_{model}}$. And $d_k = d_v = d_{model}/h$.

2) Fully connected feed-forward layer

A fully connected feed-forward layer can further improve the model on each position separately and identically. A layer consists of two linear projection with ReLU activation in between is used as

$$FFN(x) = ReLU(xW_1)W_2$$

Where $W_1, W_2$ are the weight matrices of fully connected feed-forward neural networks.

In summary, Multi-head attention in the Global Encoder projects input nodes into three different subspaces. The relationship between each pair of nodes is learned by Scaled Dot-Product Attention. No matter the distance between nodes far or close, they are treated in the same way. So, even the hidden spatial feature between two far-away nodes can be extracted. The spatial features extracted are global, and dynamically change depending on different inputs.

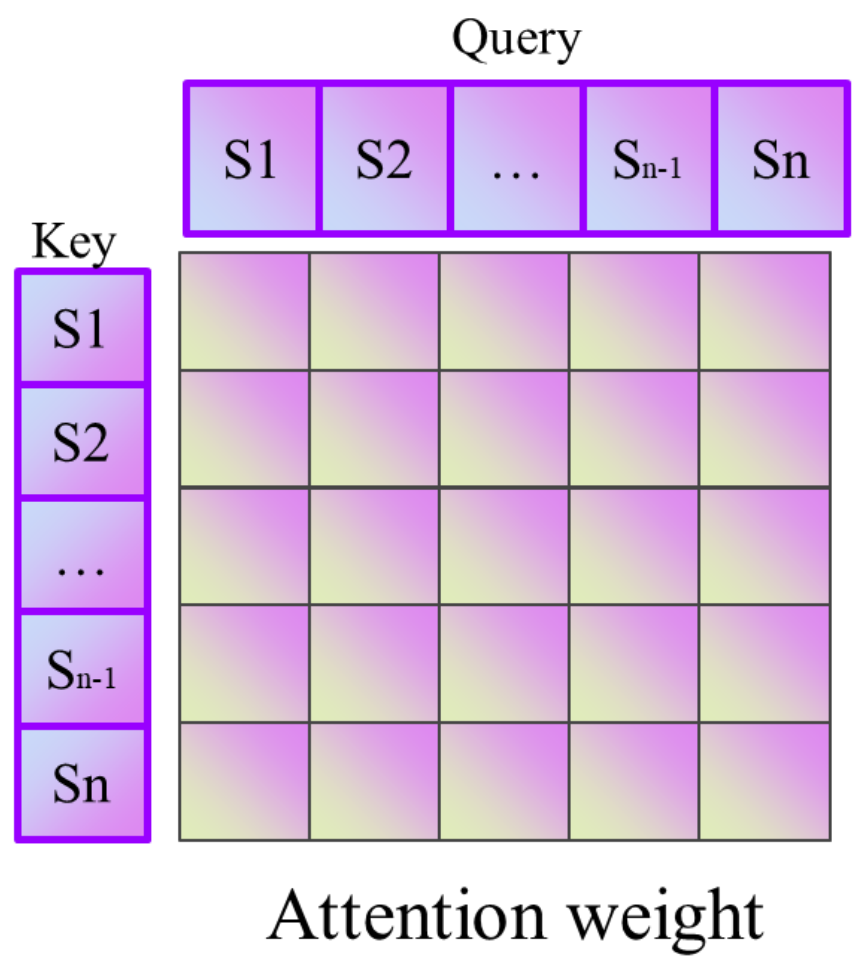

**Figure 3: Attention weight matrix comes from every pair of nodes.**

### *3.4 Global-Local Decoder*

This subsection concentrates on two things. First, it extracts the local spatial features, then fuses the local spatial features with the global features. Each Global-Local Decoder has a Mask Multi-head attention, a Multi-head attention and a feed-forward neural network. Residual connections are also added around each sub-layer, followed by layer normalization, to stabilize the gradient and help training the model better.

Mask Multi-head attention is a variant of Multi-head attention that using a mask to ignore the non-local nodes to extract local spatial features. It uses a K-hop adjacent matrix as mask to define the local and non-local, and it can be formulated as

$$\boldsymbol{M} = \sum_{1}^{K} \boldsymbol{A}$$

$$Mask(x_{ij}; \boldsymbol{M}) = \begin{cases} x_{ij} & (M_{ij} \neq 0) \\ -\infty & (M_{ij} = 0) \end{cases}$$

$$Masked_Attention(Q, K, V) = softmax\left(Mask(\frac{QK^T}{\sqrt{d_k}}; \boldsymbol{M})\right)V$$

where $\boldsymbol{M}$ represent the K-hop adjacent matrix mask, and function $Mask(\cdot)$ is added into attention weights.

The Multi-head attention in Global-Local Decoder is similar to the Multi-head attention in Global Encoder, but Global Encoder's output is used as the keys and values, i.e., $K$ and $V$, and Masked Multi-head attention's output is used as queries, i.e., $Q$. This Multi-head attention is a fusion of global and local spatial features that performs better than simply adding or concatenate by attention mechanism.

The third is a fully connected feed-forward layer that has same structure of the fully connected feed-forward layer in encoder.

In summary, the Global-Local Decoder focus on local. Although the Global Encoder should have the ability to handle the weights of every node ideally, in practice it still needs a manually defined "local" mask to learn better. And the attention-based fusion of global and local features allows the model to learn the weight of importance of these features, shows better performance in final prediction.

### *3.5 Positional Encodings and Embeddings*

The above showing structures of Traffic Transformer has only feed-forward structures without convolutional or recurrent operation. Therefore, these structures can't make use of the order of sequence, but as we mentioned before, in both traffic forecasting and NLP, the absolute and relative position is important. Changing the order of the sequence will change the meaning. The order of the sequence represents the order of the nodes and the structure of the graph. Absolute and relative position is too important to lose. To this end, we add "positional encoding" and "positional embedding" to the input of Traffic Transformer.

Positional encoding use sine and cosine functions of different frequencies, sum into each node feature:

$$PE_{(pos,2i)} = \sin(pos/10000^{2i/d_{model}})$$
$$PE_{(pos,2i+1)} = \cos(pos/10000^{2i/d_{model}})$$

where $pos$ is the position and $i$ is the dimension. This "positional encoding" is fixed and it encodes the absolute positional information. We also adopt learnable spatial positional embedding to allow the model learning something more than absolute positional information. Specifically, a learnable tensor $\mathcal{P} \in \mathbb{R}^{N \times d_{model}}$ is sum into each node.

### *3.6 Temporal Embedding*

The structures and modules mentioned above ignore the temporal features. But as our problem is a time-series modeling problem, the temporal features cannot be ignored. Most proposed models stack temporal blocks and spatial blocks alternatively. So, these models extract temporal features and spatial features step by step. Different to these models, we find it is no need to stack temporal blocks. A single temporal block at the beginning of the model is well-enough.

Besides, temporal features are usually short in practice that around ten timesteps. We use Long-Short Term Memory (LSTM) to apply the temporal embedding. The reason we choose LSTM is that LSTM can be seen as a decomposition process. LSTM projects the input from $\mathbb{R}^{T \times N \times C}$ to $\mathbb{R}^{N \times C}$ that significantly improve the efficiency and shrink the memory cost. LSTM can generally be formulated as

$$f_t = \sigma(W_f\ (h_{t-1}, x_t) + b_f)$$
$$i_t = \sigma(W_i(h_{t-1}, x_t) + b_i)$$
$$\tilde{C}_t = tanh(W_C(h_{t-1}, x_t) + b_C)$$
$$C_t = f_t * C_{t-1} + i_t * \tilde{C}_t$$
$$o_t = \sigma(W_o(h_{t-1}, x_t) + b_o)$$
$$h_t = o_t * tanh(C_t)$$

where $f_t$ is the forget gate, $i_t, \tilde{C}_t$ represent the memory gate, $C_t$ is the state of cell, $o_t$ is the output gate, $h_t$ is the hidden state. $\sigma$ represents the Sigmoid function and $W, b$ represent the weights and bias respectively. Each timestep is projected in order and different hidden states is returned as $\{h_1, h_2, \ldots, h_{t-1}, h_t\}$. And only the last hidden state $h_t$ is used.

## 4 Experiments and Results

In this section, experiments on three real-world road network are conducted to answer: (1) Whether the proposed Traffic Transformer model will significantly boost the prediction performance compared with the state-of-the-art methods? (2) Whether the Traffic Transformer model will extract features dynamically and hierarchically by attention mechanism? (3) Whether the latent spatial global and local relationship will be extracted? All the experiments are conducted using a computing platform with a NVIDIA RTX 2080Ti GPU (11GB RAM). Python is used to conduct and evaluate all experiments.

### *4.1 Data description*

Our model is verified on three traffic datasets. One of them (METR-LA) is public dataset released by previous work, and the other (Urban-BJ, Ring-BJ) are generated by ourselves.

Specifically, METR-LA is collected form loop detectors in the freeways of Los Angeles County. And this dataset ranges from Mar 2012 to Jun 2012, and there are 207 sensors(nodes) in total. Urban-BJ and Ring-BJ are generated from the uploaded GPS data of more than 10,000 taxis in Beijing. The time range of Urban-BJ and Ring-BJ are from June 2015 to August 2015. In Urban-BJ, total 278 nodes in a downtown area are selected. And in Ring-BJ 236 nodes in the second ring road of Beijing are selected. These nodes and road networks are shown in Fig.4. As shown in Table.1, METR-LA only consists of freeways and Ring-BJ only consist of primary class road. But Urban-BJ consist of different class of roads. METR-LA has a large area scope, Ring-BJ in the middle, and Urban-BJ in a small area scope. As roads of different class mixing in a small area, Urban-BJ shows a complex road network. Ring-BJ's network is simple as it just includes a ring road. The complexity of METR-LA is in the middle. We select these datasets in different characteristics to test Traffic Transformer's performance on different situations.

For all of these spatiotemporal datasets, data is aggregated every 5 minutes, observation window is 60 minutes and the maximum prediction horizon is 60 minutes. So, the length of input and output sequences in time-dimension are both 12, i.e., $\boldsymbol{X}_{n\times N\times C} = \mathcal{F}(\boldsymbol{X}_{m\times N\times C}; \mathcal{G})$ $(m = n = 12)$. The traffic speeds are normalized with Z-Score. The perioral 70% data is used for training, the next 10% data is used for validation, and the final 20% data is used for testing.

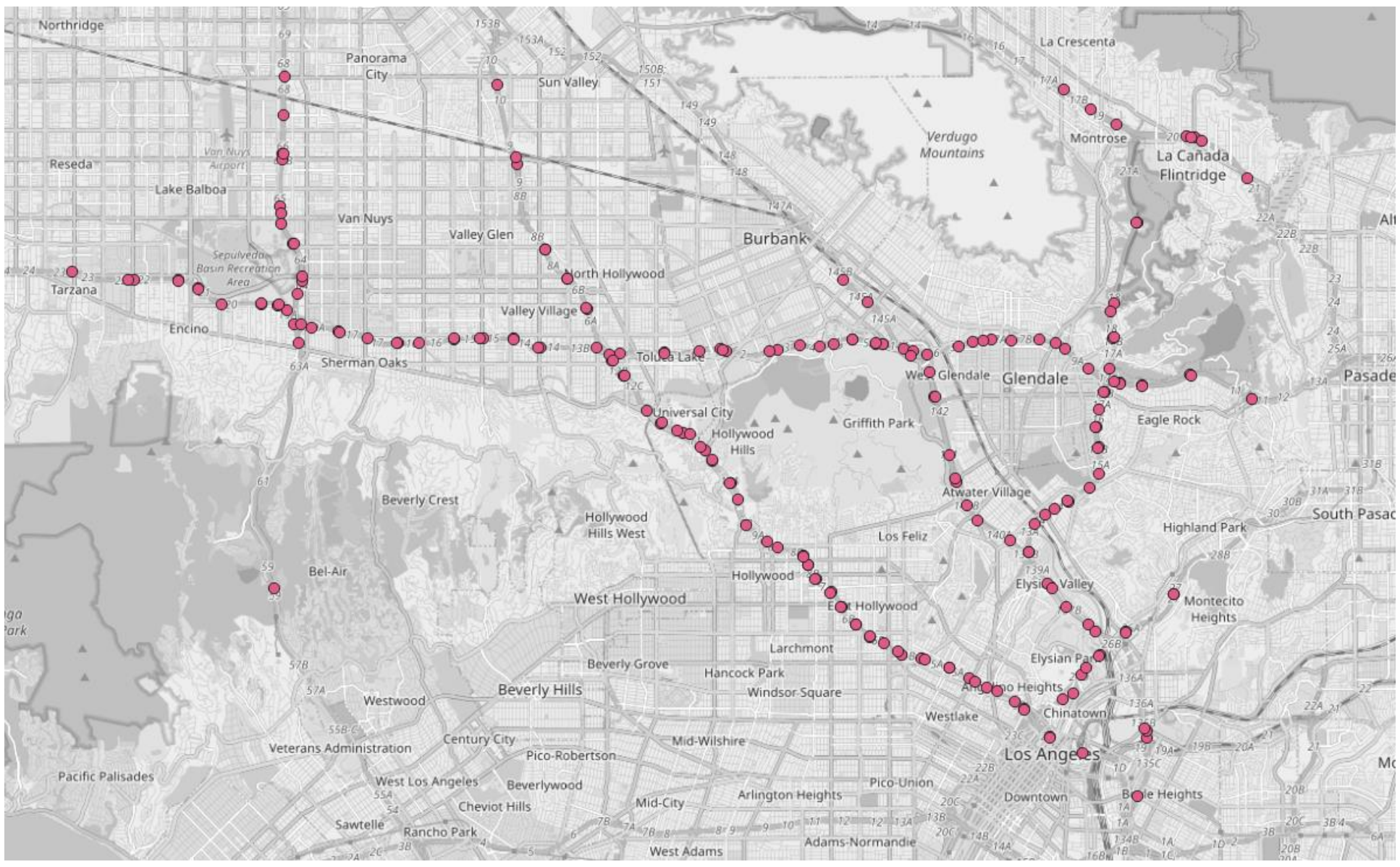

(a)

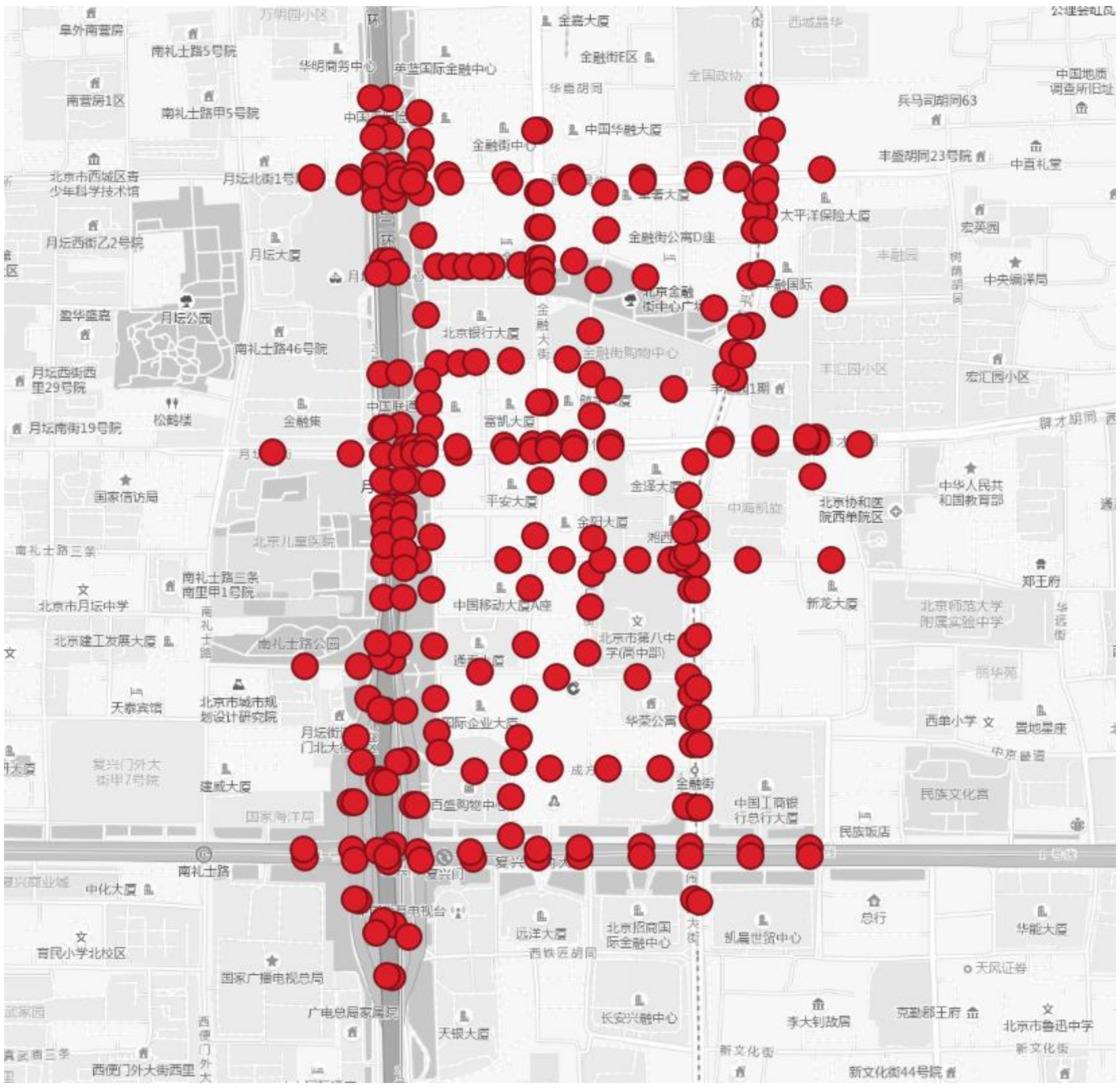

(b)

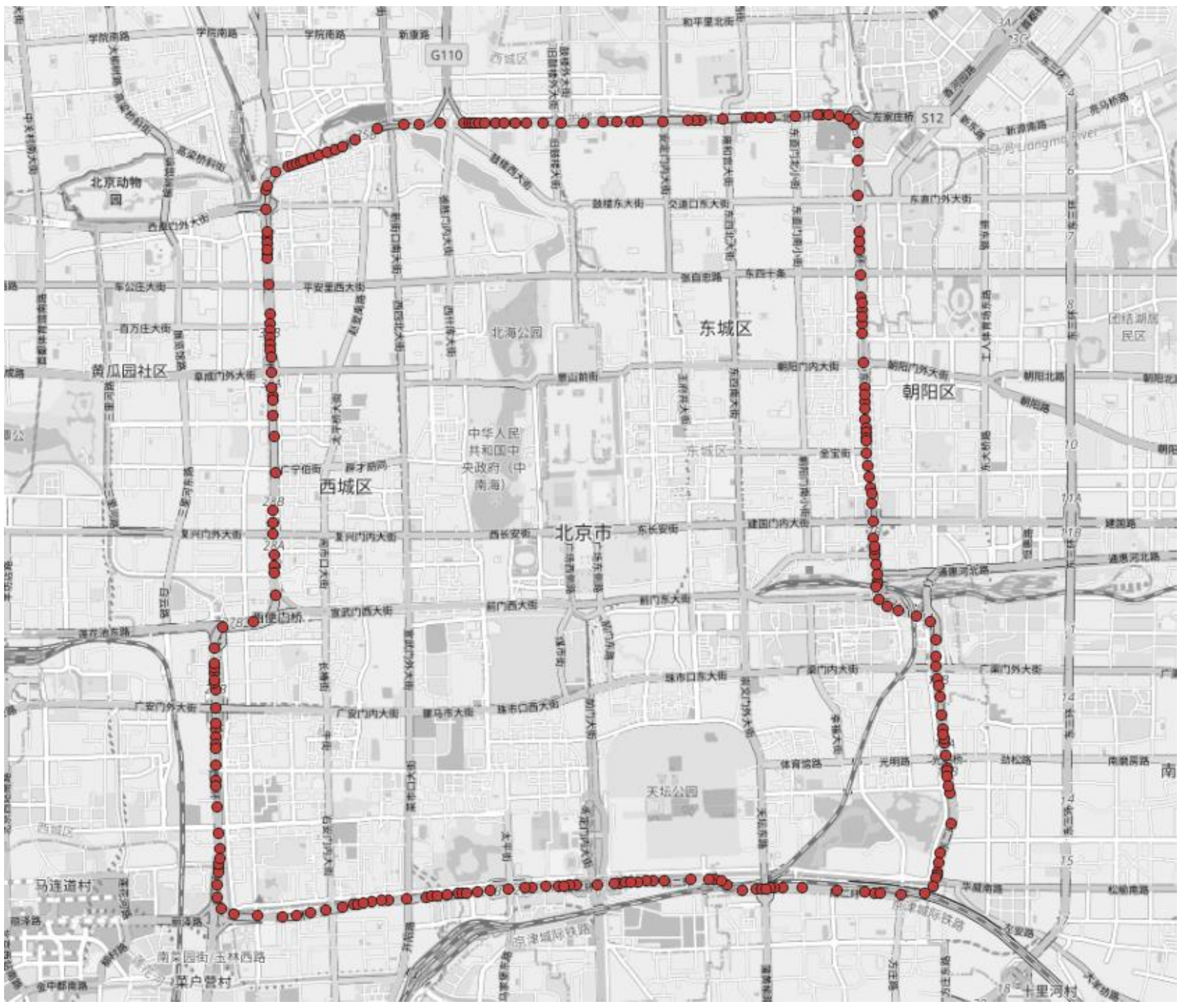

(c)

**Figure 4: The traffic networks. (a) is METR-LA, (b) is Urban-BJ, (c) is Ring-BJ.**

**Table 1: Comparison of three datasets:**

| Dataset | Road Class | Area Scope | Network complexity | Number of nodes |
|---|---|---|---|---|
| METR-LA | Freeway | Large | Middle | 207 |
| Urban-BJ | Mixed | Small | Complex | 278 |
| Ring-BJ | Primary | Middle | Simple | 236 |

### *4.2 Measures of effectiveness*

In this study, we use three measures of effectiveness, mean absolute error (MAE), mean absolute percentage error (MAPE), and root mean square error (RMSE), to evaluate the accuracy of the forecasting models:

$$MAE = \frac{1}{Nn}\sum_{k=1}^{N}\sum_{t=1}^{n}\left|\left(v_{t,k} - \hat{v}_{t,k}\right)\right|$$

$$MAPE = \frac{1}{Nn}\sum_{k=1}^{N}\sum_{t=1}^{n}\frac{\left|\left(v_{t,k} - \hat{v}_{t,k}\right)\right|}{\left|\left(v_{t,k}\right)\right|} \times 100\%$$

$$RMSE = \sqrt{\frac{1}{Nn}\sum_{k=1}^{N}\sum_{t=1}^{n}\left(v_{t,k} - \hat{v}_{t,k}\right)^2}$$

where $v_{t,k}$ is the measured traffic speed and $\hat{v}_{t,k}$ is the predicted traffic speed of timestep $t$ and node $k$. Among these different measures of effectiveness, the MAE measures the overall accuracy; the MAPE is particularly sensitive to the error in the patterns of low-speed and measures whether the model track the congestions in the road networks; and the RMSE measures both bias and variance for the uncertainty in the prediction.

### *4.3 Model comparisons*

Our proposed Traffic Transformer model is compared with the following selected five benchmark models, which contain the traditional models, including ARIMA and FC-LSTM, and GCN-based models, including DCRNN, STGCN and GWN.

- ARIMA: A popular model using in time series prediction. The orders of autoregression, difference, and moving average are the three critical parameters for the ARIMA model. The optimal parameters are obtained from grid search by "auto-ARIMA" tools.
- FC-LSTM: Fully-Connected LSTM (FC-LSTM) can extract long-short term dependencies on temporal features, which is a classic RNN to learn time series and make prediction by fully connected neural network.
- DCRNN: Diffusion Convolutional Recurrent Neural Network (DCRNN) is one of the most representative GNN methods for traffic forecasting. It proposes a sequence to sequence (seq2seq) model with encoder-decoder structure to predict in multi-steps.
- STGCN: Spatio-Temporal Graph Convolutional Network (STGCN) is another the most representative GNN methods for traffic forecasting. It uses the stack of temporal block and spatial block to extract ST-features and predict in single-step.
- GWN: Graph Wave Net (GWN) adapt Wave Net model to Graph problem and has achieved the state-of-the-art performance.

In addition, to further demonstrate the effectiveness of different parts of Traffic Transformer, we design an ablation study that uses three variants to compared with Traffic Transformer:

- Traffic Transformer Encoder: only the Global Encoder is using in Transformer structure, the Global-Local Decoder is completely removed. So, this model trained without predefined adjacent matrix. The graph structure is completely learned by model.
- Traffic Transformer Decoder: only the Global-Local Decoder is using in Transformer structure, the Global Encoder is completely removed. As the Global Encoder is missing, the second multi-head attention layer of Global-Local Decoder is not established, and it is removed. So, this model can also be seen as a Global Encoder with a mask.
- Traffic Transformer-No temporal: the temporal embedding at the beginning of the model is removed. This model can’t extract temporal features of data. The speeds at different timesteps are flatten in one dimension and projected to original dimension by linear neural network.

Grid search is used to tuning up the hyperparameters. Finally, we set the number of Global Encoder block and Global-Local Decoder block both to 6. The hidden feature channel in our model after temporal embedding is set to 64. The dim of feed-forward network in Traffic Transformer is set to 256. We train our models by minimizing MAE and use Adam as optimizer with 1e-5 weight decay. Early stopping on the validation set is used to mitigate overfitting. Clipping gradient mechanism is used to stabilize the gradient to train better.

The Table 1 to Table 3 list the results in different terms of Traffic Transformer and the compared benchmark methods on METR-LA, Urban-BJ and Ring-BJ respectively. The results indicate Traffic Transformer and Traffic Transformer Encoder achieve the best performance on three datasets and different terms of predictions (However, GWN is better in the comparison of RMSE in short-term (15min) and middle-term (30min) on Ring-BJ), especially in long-term (60min) prediction, because Traffic Transformer can benefit from the hidden long-distance spatial relationships extracted. Some observations and phenomena can be gotten by further analyses.

- In the results of traditional models, FC-LSTM performs much better than ARIMA that indicates the deep learning methods is more suitable for traffic forecasting.
- GCN-based models show significantly higher accuracy than traditional models. These GCN-based models indicate spatial features are irreplaceable in traffic forecasting. In other words, spatiotemporal models are much better performance than temporal-only models.
- GWN, Traffic Transformer and Traffic Transformer Encoder achieves much better results than DCRNN and STGCN, because GWN can learn adaptive adjacent matrix once per iteration, and Traffic Transformer and Traffic Transformer Encoder can give adjacent relationship of every pair of nodes through attention mechanism. So, GWN, Traffic Transformer and Traffic Transformer Encoder can extract hidden spatial features allowing better performance especially in long-term.
- Compared to GWN, in our proposed Traffic Transformer and Traffic Transformer Encoder, the learnt spatial relationships come from Multi-head attention mechanism, so the learnt spatial relationships is dynamic and depending on input data. And the deep structure of Traffic Transformer and Traffic Transformer Encoder extract hierarchical spatiotemporal features rather than a single adjacent matrix. Therefore, our models extract hierarchical hidden spatial features dynamically on different situations, and performs better than GWN, especially in long-term prediction.
- In Urban-BJ, the ablation model Traffic Transformer Encoder is better than origin model and achieve the best result. It is because a bad-defined adjacent matrix would harm the model. And Urban-BJ has high complexity shown in Table 1 that it is difficult to define a well-enough adjacent matrix. Therefore, for the similar situation, it is better to use Traffic Transformer Encoder only rather than Traffic Transformer.
- Even without temporal embedding block, Traffic Transformer No temporal's result is not bad. It indicates that the temporal features hidden in the data is simple, the changes in time-dimension is no need to extract again and again. Only a single temporal embedding at the beginning is enough and significantly shrink the training cost as a decomposition.

We also compare the computation cost of Traffic Transformer with DCRNN, STGCN and GWN on the largest dataset, Urban BJ. As shown in Table 3, DCRNN is much slower than other models because DCRNN's multi-step prediction architecture is too expensive in training. Our Traffic Transformer is faster than DCRNN but slower than STGCN and GWN in training. In testing, the one-step prediction models are significantly efficient and Traffic Transformer are faster than DCRNN but slower than STGCN and GWN.

**Table 1: Comparison of performance on METR-LA**

| Model | 15min | | | 30min | | | 60min | | |
|---|---|---|---|---|---|---|---|---|---|
| | MAE | RMSE | MAPE | MAE | RMSE | MAPE | MAE | RMSE | MAPE |
| ARIMA | 3.99 | 8.21 | 9.60% | 5.15 | 10.45 | 12.70% | 6.9 | 13.23 | 17.40% |
| FC-LSTM | 3.44 | 6.3 | 9.60% | 3.77 | 7.23 | 10.90% | 4.37 | 8.69 | 13.20% |
| DCRNN | 2.77 | 5.38 | 7.30% | 3.15 | 6.45 | 8.80% | 3.6 | 7.6 | 10.50% |
| STGCN | 2.88 | 5.74 | 7.62% | 3.47 | 7.24 | 9.57% | 4.59 | 9.4 | 12.70% |
| GWN | 2.69 | 5.15 | 6.90% | 3.07 | 6.22 | 8.37% | 3.53 | 7.37 | 10.01% |
| **Traffic Transformer** | **2.66** | **5.11** | **6.75%** | **3.00** | **6.06** | **8.00%** | **3.39** | **7.04** | **9.37%** |
| Traffic Transformer Encoder | 2.69 | 5.23 | 6.89% | 3.03 | 6.17 | 8.12% | 3.43 | 7.22 | 9.54% |
| **Traffic Transformer Decoder** | **2.66** | **5.11** | 6.81% | 3.03 | 6.13 | 8.15% | 3.49 | 7.31 | 9.82% |
| Traffic Transformer No temporal | 2.75 | 5.28 | 7.27% | 3.12 | 6.32 | 8.66% | 3.52 | 7.35 | 10.12% |

## Table 2: Comparison of performance on Urban-BJ

| Model | 15min | | | 30min | | | 60min | | |
|---|---|---|---|---|---|---|---|---|---|
| | MAE | RMSE | MAPE | MAE | RMSE | MAPE | MAE | RMSE | MAPE |
| ARIMA | 6.13 | 8.29 | 24.22% | 6.09 | 8.20 | 23.6% | 6.15 | 8.28 | 23.78% |
| FC-LSTM | 4.50 | 6.59 | 17.29% | 4.85 | 6.96 | 19.06% | 5.08 | 7.23 | 20.23% |
| DCRNN | 4.38 | 6.45 | 16.72% | 4.70 | 6.81 | 18.29% | 4.90 | 7.08 | 19.19% |
| STGCN | 4.49 | 6.52 | 17.37% | 4.74 | 6.80 | 18.66% | 4.91 | 7.03 | 19.61% |
| GWN | 4.56 | 6.58 | 18.01% | 4.71 | 6.81 | 18.77% | 4.84 | 6.99 | 19.30% |
| **Traffic Transformer** | 4.34 | 6.40 | 16.67% | 4.62 | 6.70 | 18.15% | **4.77** | **6.91** | **18.89%** |
| **Traffic Transformer Encoder** | **4.32** | **6.34** | **16.65%** | **4.60** | **6.69** | **18.08%** | **4.77** | **6.91** | 18.91% |
| **Traffic Transformer Decoder** | 4.36 | 6.40 | 16.69% | 4.65 | 6.75 | **18.08%** | 4.83 | 6.99 | 19.06% |
| Traffic Transformer No temporal | 4.43 | 6.45 | 16.94% | 4.71 | 6.81 | 18.29% | 4.88 | 7.04 | 19.17% |

## Table 3: Comparison of performance on Ring-BJ

| Model | 15min | | | 30min | | | 60min | | |
|---|---|---|---|---|---|---|---|---|---|
| | MAE | RMSE | MAPE | MAE | RMSE | MAPE | MAE | RMSE | MAPE |
| ARIMA | 4.16 | 6.97 | 12.38% | 4.27 | 7.29 | 12.77% | 4.22 | 7.20 | 12.84% |
| FC-LSTM | 2.41 | 4.44 | 6.39% | 2.91 | 5.54 | 8.32% | 3.54 | 6.80 | 11.02% |
| DCRNN | 2.38 | 4.23 | 6.21% | 2.86 | 5.34 | 7.89% | 3.47 | 6.65 | 10.14% |
| STGCN | 2.47 | 4.46 | 6.55% | 2.94 | 5.46 | 8.27% | 3.56 | 6.64 | 10.45% |
| **GWN** | 2.30 | **4.07** | 5.97% | 2.75 | **5.06** | 7.65% | 3.39 | 6.26 | 10.03% |
| **Traffic Transformer** | 2.31 | 4.15 | 6.08% | **2.71** | 5.13 | 7.63% | **3.22** | **6.25** | **9.69%** |
| **Traffic Transformer Encoder** | **2.28** | 4.10 | **5.94%** | 2.73 | 5.10 | **7.60%** | 3.33 | 6.36 | 9.85% |
| Traffic Transformer | 2.37 | 4.25 | 6.12% | 2.82 | 5.30 | 7.81% | 3.40 | 6.52 | 10.05% |
| Decoder | | | | | | | | | |
| Traffic Transformer No temporal | 2.42 | 4.24 | 6.32% | 2.82 | 5.25 | 7.94% | 3.35 | 6.38 | 10.12% |

**Table 4: Comparison of the computation cost on BJR dataset**

| Model | Computation Time | |
|---|---|---|
| | Training(s/epoch) | Testing(s) |
| DCRNN | 814.27 | 57.66 |
| STGCN | 40.46 | 1.57 |
| GWN | 98.21 | 4.36 |
| Traffic Transformer | 140.96 | 5.15 |

### *4.4 Model Interpretation*

Our experiment indicates that models with learnable adjacent matrix, in other words, models can learn spatial relationships perform much better, especially in long-term predictions. Traffic Transformer learns spatial relationships through Multi-head attention in both Global Encoder and Global-Local Decoder. The dynamic and hierarchical attention weight matrixes of each multi-head attention are similar to adjacent matrix and can reveal the learned spatial relationships.

We design another experiment to better demonstrate the dynamic and hierarchical spatiotemporal features Traffic Transformer learned. Some batches of test data are chosen and are projected to the trained model, and the attention weight matrixes are returned for analyze. For example, we choose three batches in the test set of METR-LA in three different time periods: 0:00 AM, 8:00AM, 4:00PM. And these batches are projected into our trained model separately. The attention weight matrixes of each multi-head attention layer and each dataset are scaled into $(0, 1)$ and are given in Fig.5.The attention weight matrixes in Global encoder is called "source weight" and are abbreviated to "src" in figure, the first attention weight matrixes in Global-Local decoder is called "target weight" and are abbreviated to "tgt" in figure, and the second attention weight matrixes in Global-Local decoder is called "memory weight" and are abbreviated to "mem" in figure.

In each attention weight heatmap, the weight of $i^{th}$ row and $j^{th}$ column represents the impact of $sensor_i$ to $sensor_j$. As shown in Fig.5, the weights of some rows and columns are much higher than others, it means that the nodes corresponding to these rows and columns are important to almost all the other nodes. It suggests that these nodes are influential to most nodes in the graph while other nodes impact weaker. The source attention weights differ in different time periods and layers, which means Traffic Transformer can dynamically and hierarchically extract the global spatial features base on input data. The target attention weights look very similar. It is because the target attention weights using the predefined K-hop adjacent matrix to focus on local. The memory attention weights show how global and local features are fused. Their graphs show similar characteristics as source attention weight heatmaps that some nodes are influential to most nodes in the graph in fusion, while others aren't.

To further demonstrate our observation, we define the importance of each node through

$$I_k = \sum A_{ik} + \sum A_{kj}$$

where $I_k$ represents the importance of the node $k$, and $A$ represents the attention weight matrix. And the node whose value is higher than one standard deviation to the mean values is defined as the influential nodes.

1) Dynamicity

We choose two batches of data in the test set of METR-LA. One is on the time period of 8:00AM, and another is on the time period of 4:00PM. These two batches are sent to the trained model, and the attention weight matrixes in all Multi-head attention blocks are returned with the output. We choose the attention weight matrixes of the "memory 6" (i.e. mem 6) for further analysis. Sensor NO.144 is chosen to analyze. The top 10 influential sensors of sensor NO.144 are selected and plot on Fig.6. As shown on Fig.6, the red dot represents sensor NO.144. The blue dots represent the top 10 influential sensors on the time period of 8:00 AM and the orange dots represent the top 10 influential sensors on the time period of 4:00PM. The blue dots are placed on the eastern of the network and the red dots are place on the western of the network. To be point out, the sensor NO.162 and the sensor NO.163 are close in distance, but they are in the different directions of the same freeway section as shown in the enlarged part of Fig.6. It means that Traffic Transformer finds different influential points in the different time period. It indicates that our propose Traffic Transformer can extract spatial dependencies dynamically, the adjacent relationship in our model is dynamic and depending on inputs. It can break through the limitation that the adjacent matrix is fixed in the traditional GCN-based models, and perform better.

2) Hierarchy

In METR-LA dataset, we choose a batch of 8:00AM in test set. We calculate the influential nodes of different source attention layers and plot them on map as shown on Fig.7. It can be seen that most of the influential nodes locate nearby the intersections that is usually important in traffic network. The spatial features in different layers shown hierarchical characteristics that the model focus on from global to local and finally back to global. In bottom layer (Fig.7 (a), layer 1), the influential nodes distribute in whole network. Bu t in the middle layer (Fig.7 (b) and (c), layer 2 and 4), the influential nodes are concentrated on a part of the network. For example, most cyan points shown in Fig.7 (b) representing the influential nodes of layer 2 are focusing on the east part of the network. And in higher layer (Fig.7 (d), layer 6), the influential nodes distribute in whole network again. To be point out, it is interesting that in layer 6, sensor 26 (on the left side of the figure) is extracted as an influential node. However, in METR-LA point 26 is an isolated vertex. It confirms Traffic Transformer can extract long-distance spatial features.

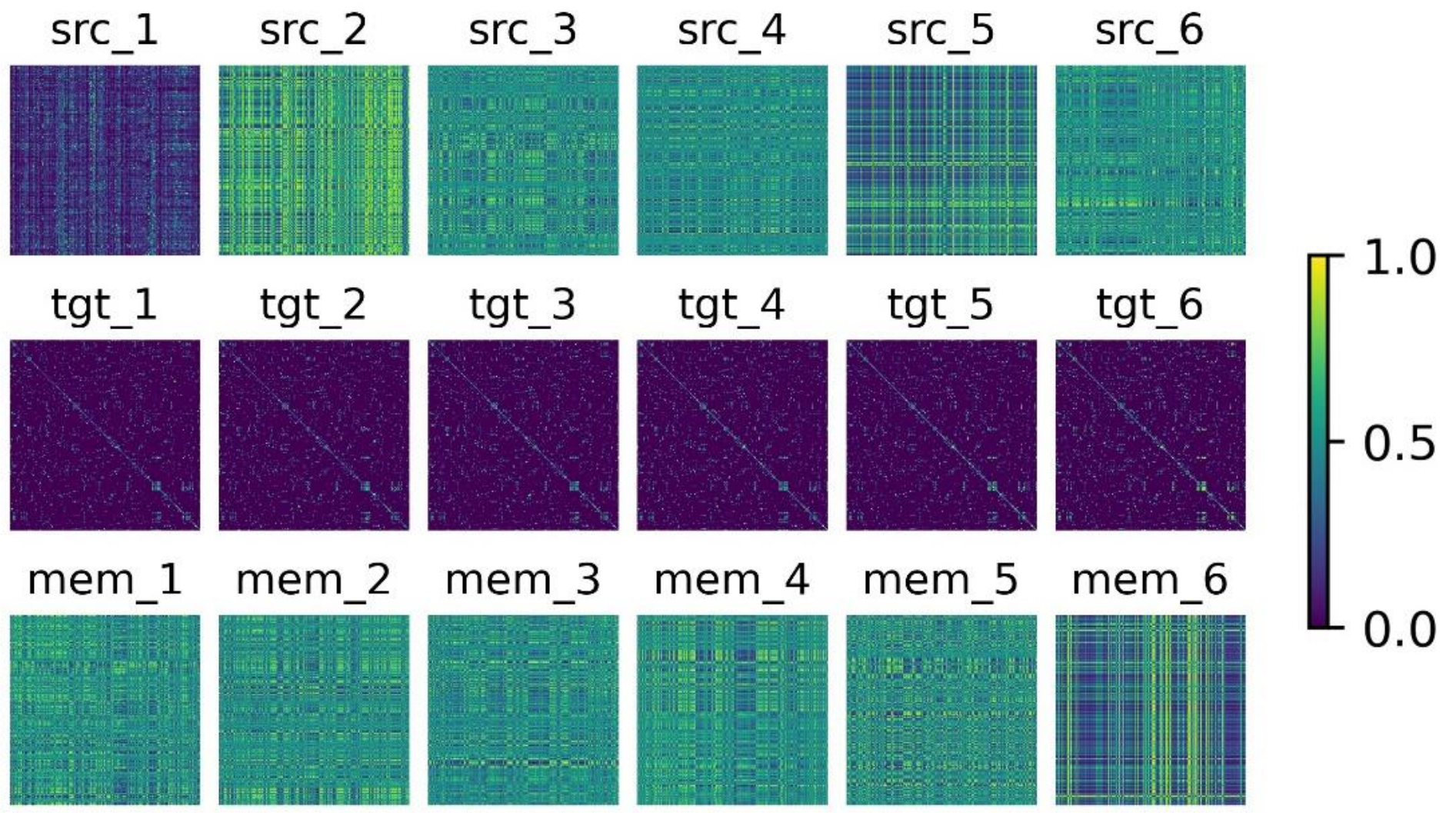
src_1 src_2 src_3 src_4 src_5 src_6
tgt_1 tgt_2 tgt_3 tgt_4 tgt_5 tgt_6
mem_1 mem_2 mem_3 mem_4 mem_5 mem_6
1.0
0.5
0.0

(a)

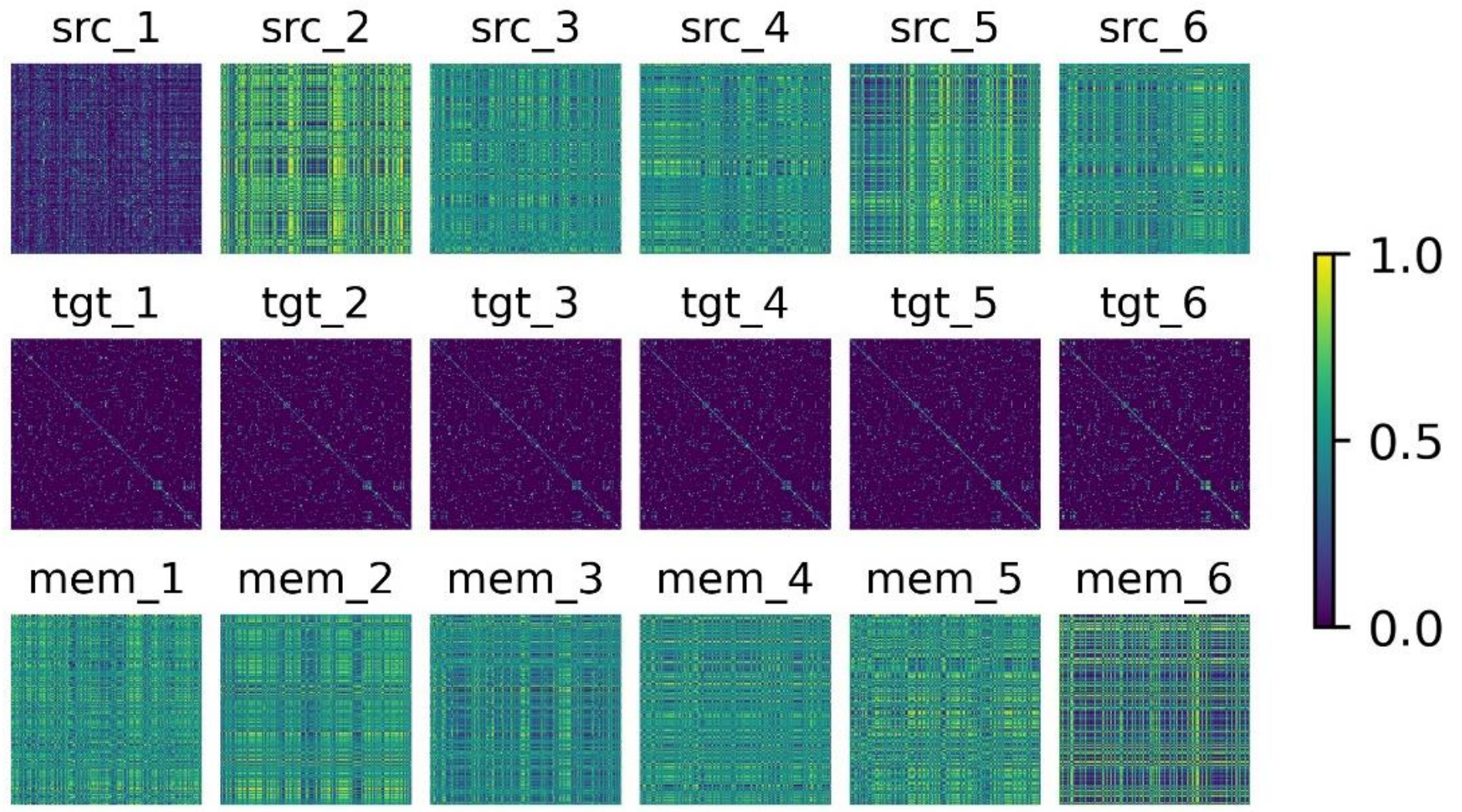
src_1 src_2 src_3 src_4 src_5 src_6
tgt_1 tgt_2 tgt_3 tgt_4 tgt_5 tgt_6
mem_1 mem_2 mem_3 mem_4 mem_5 mem_6
1.0
0.5
0.0

(b)

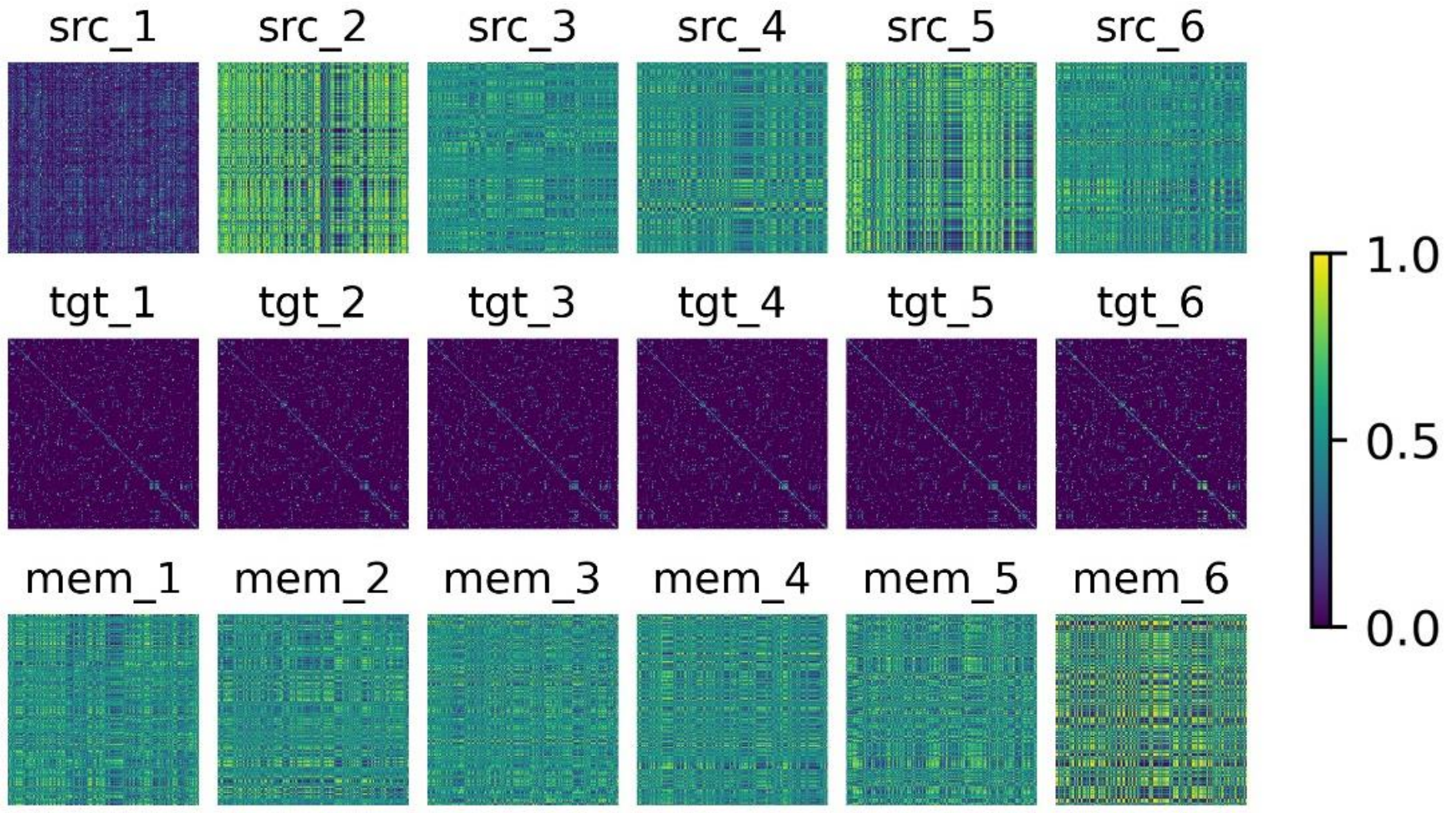
src_1 src_2 src_3 src_4 src_5 src_6
tgt_1 tgt_2 tgt_3 tgt_4 tgt_5 tgt_6
mem_1 mem_2 mem_3 mem_4 mem_5 mem_6
1.0
0.5
0.0

(c)

**Figure 5: Attention weight matrixes of different time periods. (a): Attention weight matrixes of the batch of 0:00 AM; (b): Attention weight matrixes of the batch of 8:00 AM; (c) Attention weight matrixes of the batch of 4:00 PM.**

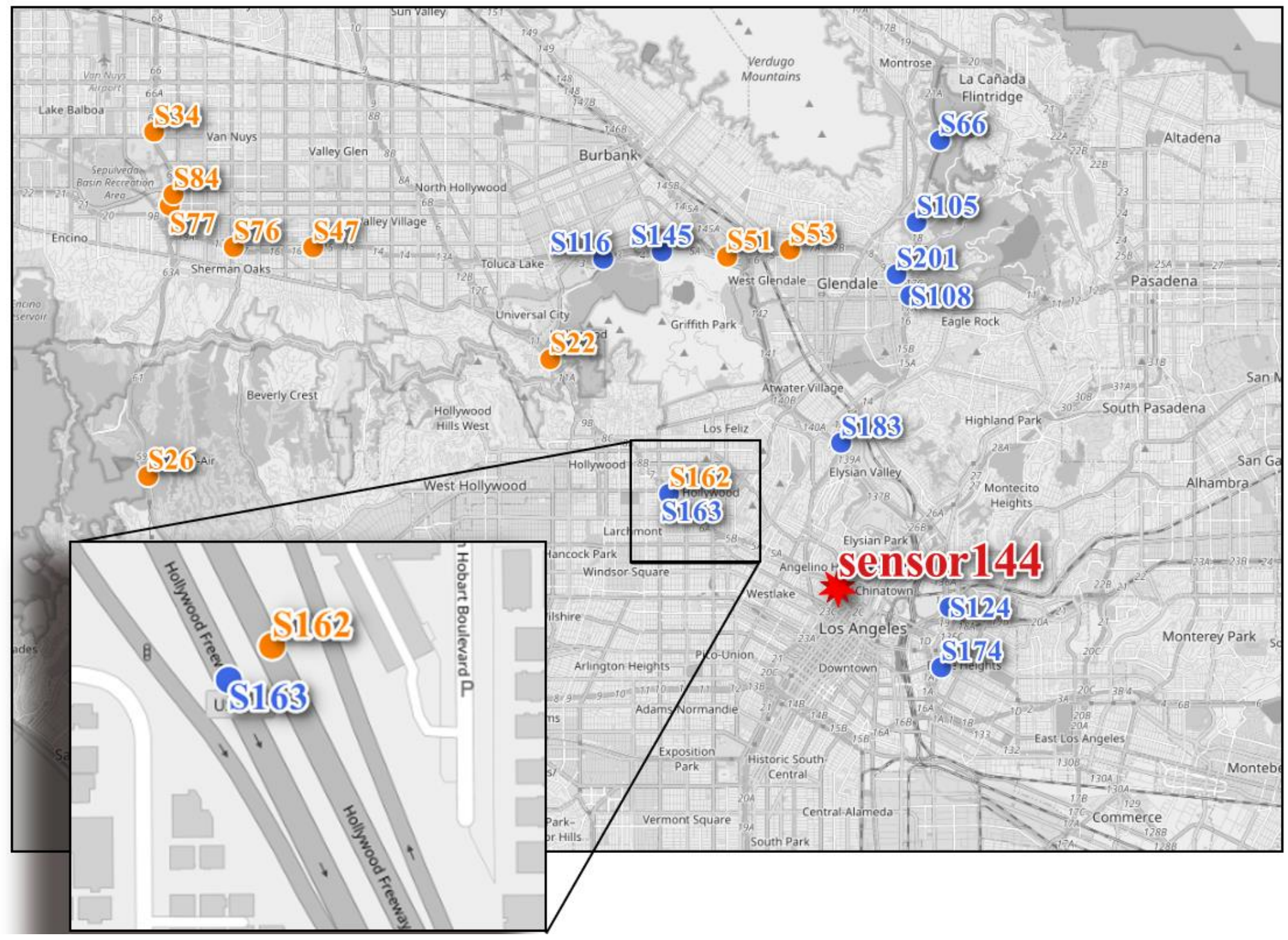

**Figure 6: Influential sensors of sensor NO.144 in different time period. Blue dots: 8:00 AM; Orange dots: 4:00 PM. The enlarged part of the picture shows two sensors in different directions on the same road.**

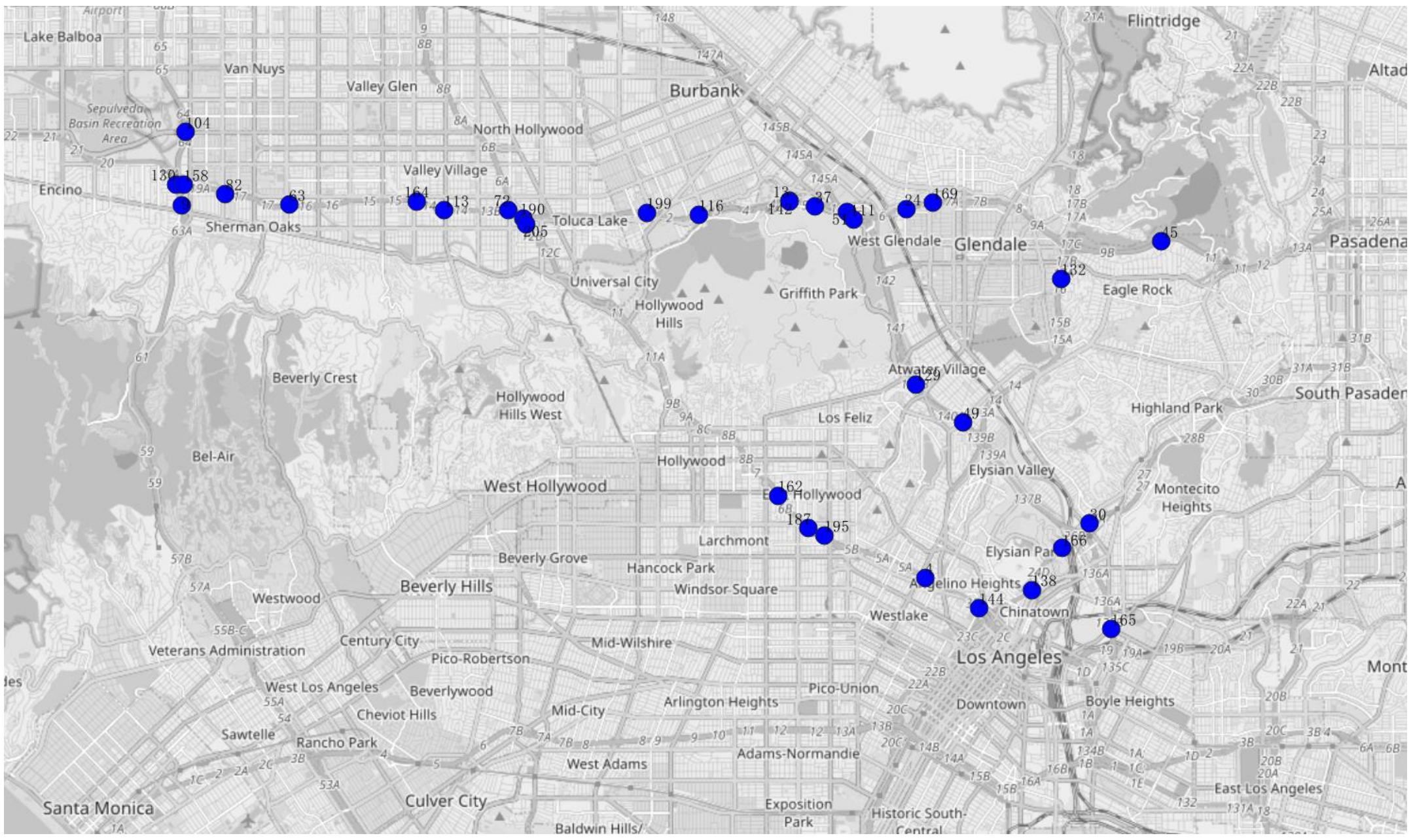

(a)

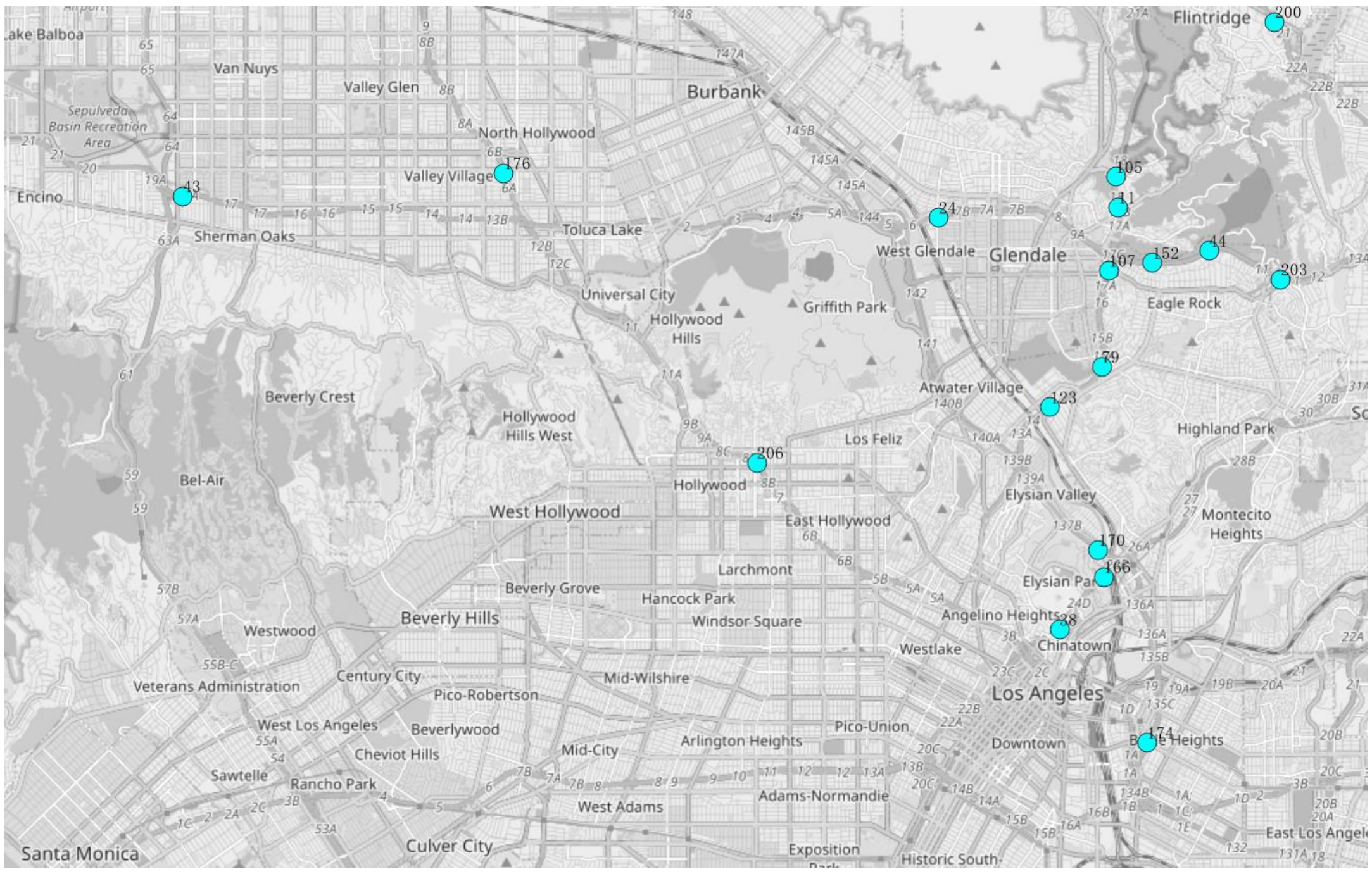

(b)

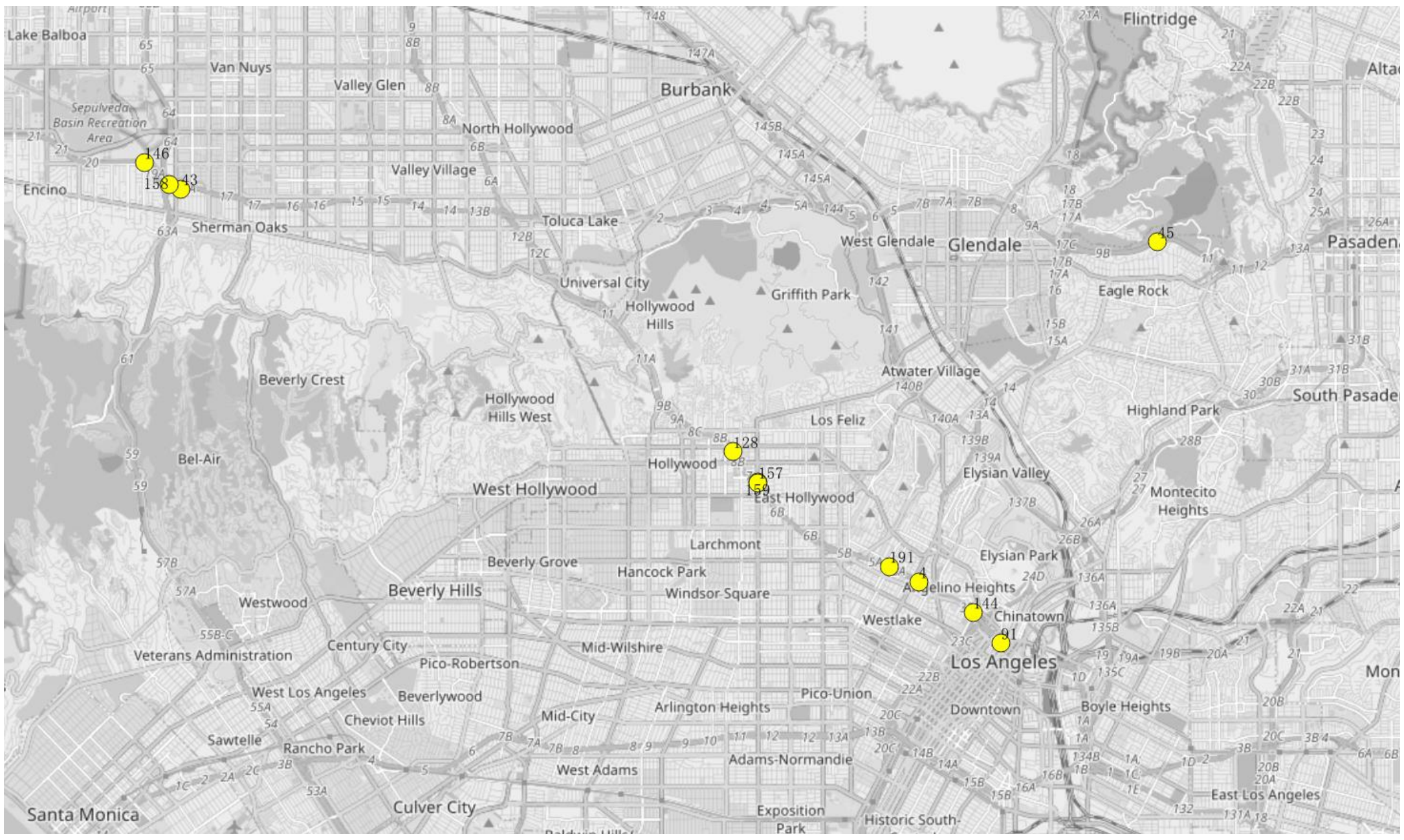

(c)

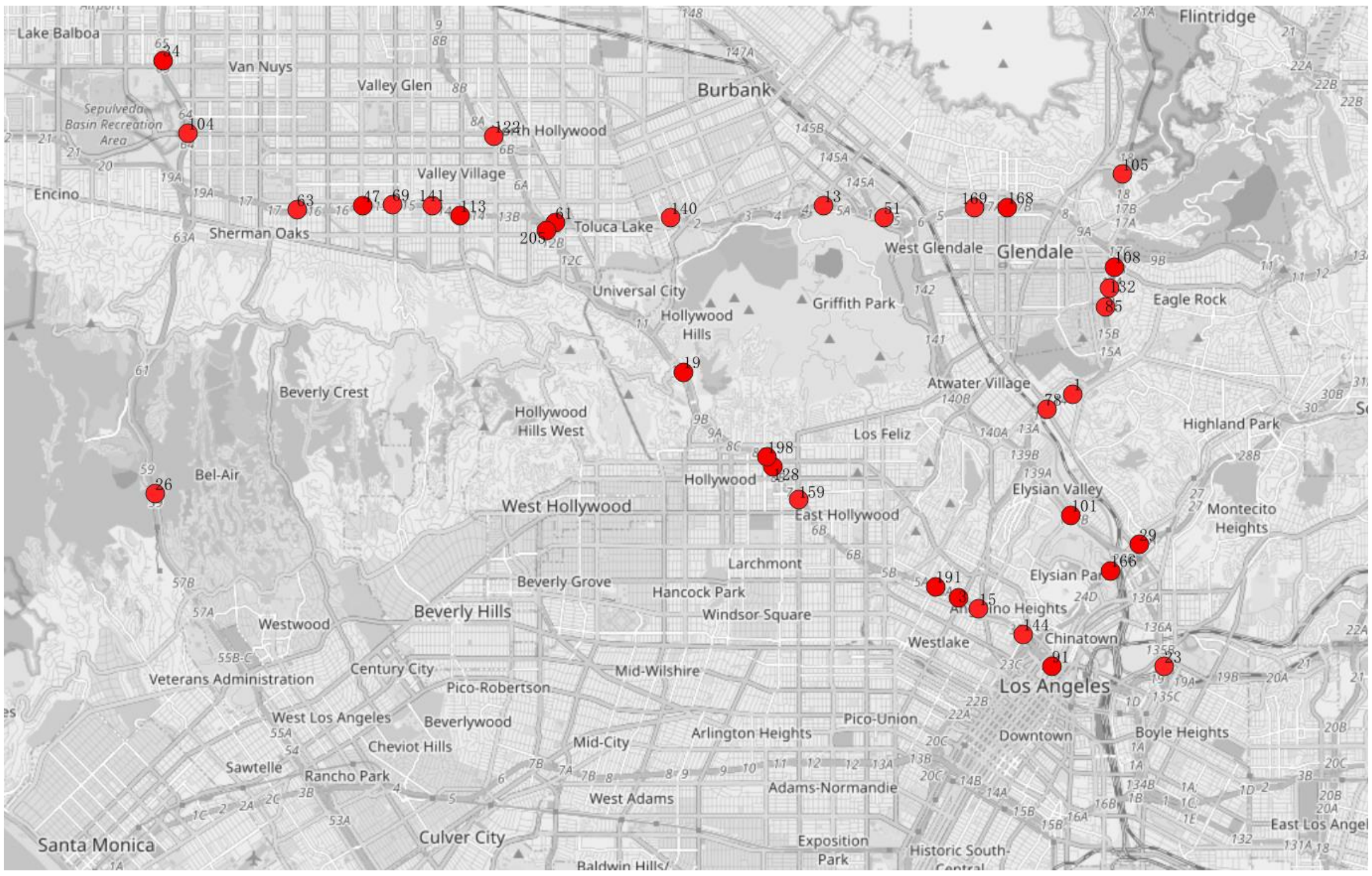

(d)

**Figure 7: Influential nodes of different layers shown on map: (a) layer 1; (b) layer 2; (c) layer 4; (d) layer 6. (a) is the bottom layer; (b) and (c) are the middle layer; and (d) is the top layer. (a) and (d) prefer global features but (b) and (c) prefer local features.**

## 5 Conclusions and Discussions

In this paper, we present a novel model for spatial-temporal graph modeling and long-term traffic forecasting. Our proposed Traffic Transformer model can efficiently and effectively extract dynamic and hierarchical traffic spatiotemporal features through data. We choose three different datasets to evaluate whether our model is robust in different situations. The proposed model is compared with eight benchmark models including traditional models, GCN-based models and ablation models. Three measures of effectiveness (i.e. MAE, MAPE and RMSE) are used to evaluate the accuracy of these models. On three real-world dataset, Traffic Transformer achieves state-of-the-art results. and helps people find the hierarchical influential nodes of network.

According to the experiment results, we can draw the following conclusions: (a) The network-wide traffic forecasting has a similar characteristic and challenge to NLP. And the advanced methods and frameworks of NLP such as Transformer can adapt to network-wide traffic forecasting problems. (b) Among all the traffic forecasting models, Traffic Transformer achieves the best prediction results within acceptable training time. Thus, it proves Transformer is powerful to traffic forecasting problem. (c) The influential sensors extracted by trained Traffic Transformer model are not in local. It indicates the spatial dependencies in road network may far in distance, and the traditional distance-based adjacent matrix is limited. (d) Different time periods having different influential sensors means that the spatial dependencies in road network are dynamic, and the traditional fixed adjacent matrix is limited. (d) The spatial dependencies are hierarchical and a deep model can extract these features in global or local. (e) As the complexity of network raising, it is harder and harder to define a well-enough adjacent matrix, and it is more unavoidable the mistake in the human-defined matrix harms the model. Let the model learn the spatial dependencies is better to improve the performance in these situations. However, if the complexity of network is moderate, using a predefined adjacent matrix to help the model better knowing the "local" can improve performance.

In future work, the proposed model can be extended to consider more external information (e.g., road features, adverse weather conditions and big events) to improve prediction accuracy. Due to the data limitation, this paper only deals with the speed forecasting in road networks, but the model can extend to other traffic states (e.g., quantity of flow and density) and other transportation systems (e.g., metro network and non-motorized traffic system) once more informative data are available. In this paper, the number of sensors is limited below 300. It is because Transformer has $O(n^2)$ algorithm complexity that the model will be hard to train when the number of sensors is too large. However, some variants of Transformer proposed recently such as Transformer-XL (Dai et.al., 2019) and Longformer (Beltagy et.al., 2020) can handle the 10,000-level sequence such as the long-document. We will study using these variants of Transformer to deal with bigger networks and find the deeper hidden dependencies within traffic. Besides, language models based on Transformer are popular in NLP such as BERT (Devlin et.al 2018) and GPT (Radford et.al., 2017; Radford et.al., 2018). Researchers consider BERT is an auto-encode (AE) procedure, and GPT is an auto-regressive (AR) procedure. In the extraction of the spatiotemporal features of traffic, the spatial dimension is more likely an AE procedure, and the temporal dimension is more likely an AR procedure. In future research, a “language model” of traffic may show tremendous performance.